\documentclass[review]{elsarticle}

\usepackage{bm}
\usepackage{lineno,hyperref}
\usepackage{mathtools}
\usepackage{graphicx}
\usepackage{multirow}
 \usepackage{textcomp}
 \usepackage{tikz}
 \usetikzlibrary{arrows,positioning}
\usepackage{makecell} 
\setcellgapes{5pt}

\usepackage{amsmath,amsfonts,amssymb}

\DeclareMathOperator*{\argmin}{arg\,min}

\usepackage{lineno,hyperref}
\modulolinenumbers[5]

\journal{}

\begin{document}

\fontsize{9}{8}\selectfont
\sloppy

\begin{frontmatter}

\title{Ordinal time series analysis with the R package \textit{otsfeatures}}


\author[mymainaddress]{\'Angel L\'opez-Oriona\corref{mycorrespondingauthor}}
\ead{oriona38@hotmail.com, a.oriona@udc.es}

\author[mymainaddress,mysecondaryaddress]{Jos\'e A. Vilar}
\cortext[mycorrespondingauthor]{Corresponding author}
\ead{jose.vilarf@udc.es}

\address[mymainaddress]{Research Group MODES, Research Center for Information and Communication Technologies (CITIC), University of A Coru\~na, 15071 A Coru\~na, Spain.}
\address[mysecondaryaddress]{Technological Institute for Industrial Mathematics (ITMATI), Spain.}

\begin{abstract}
	The 21st century has witnessed a growing interest in the analysis of time series data. Whereas most of the literature on the topic deals with real-valued time series, ordinal time series have typically received much less attention. However, the development of specific analytical tools for the latter objects has substantially increased in recent years. The R package \textbf{otsfeatures} attempts to provide a set of simple functions for analyzing ordinal time series. In particular, several commands allowing the extraction of well-known statistical features and the execution of inferential tasks are available for the user. The output of several functions can be employed to perform traditional machine learning tasks including clustering, classification or outlier detection. \textbf{otsfeatures} also incorporates two datasets of financial time series which were used in the literature for clustering purposes, as well as three interesting synthetic databases. The main properties of the package are described and its use is illustrated through several examples. Researchers from a broad variety of disciplines could benefit from the powerful tools provided by \textbf{otsfeatures}.
\end{abstract}


\end{frontmatter}

\section{Introduction}\label{sectionintroduction}

Time series data usually arise in a wide variety of disciplines as machine learning, biology, geology, finance and medicine, among many other fields. Typically, most of the works on the analysis of these objects have focused on real-valued time series, while the study of time series with alternative ranges has been given a limited attention. However, the latter type of time series naturally appear in several fields when attempting to analyze several phenomena. For instance, weekly counts of new infections with a specific disease in a particular place are often modeled through integer-valued time series \cite{weiss2014binomial}. In some contexts, the time series under consideration do not even take numerical values (e.g., temporal records of EEG sleep states for an infant after birth \cite{stoffer2000spectral}). A comprehensive introduction to the topic of time series with alternative ranges including classical models, recent advances, key references and specific application areas is provided by \cite{weiss2018introduction}. 

Categorical time series (CTS) are characterized by taking values on a qualitative range consisting of a finite number of categories, which is called ordinal range, if the categories exhibit a natural ordering, or nominal range, otherwise. In this paper, the specific case of an ordinal range is considered. Time series fulfilling this condition, frequently referred to as ordinal time series (OTS) pose several challenges to the statistical practitioner. Indeed, dealing with ordered qualitative outcomes implies that some classical analytic tools must be properly adapted. For instance, standard measures of location, dispersion and serial dependence can not be defined in the same manner as in the real-valued case, but the underlying ordering existing in the series range still allows for a meaningful definition of the corresponding quantities in the ordinal setting. For instance, in \cite{weiss2020distance}, a unified approach based on expected distances is proposed to obtain well-interpretable statistical measures for ordinal series. In addition, sample counterparts of the corresponding measures are introduced and their asymptotic properties are derived. 

Ordinal series arise in multiple fields. Some interesting examples include credit ratings of different countries \cite{weiss2020distance} or degree of cloud coverage in different regions \cite{weiss2020regime}. Such types of time series are naturally considered as ordinal. On the contrary, in many situations, the series under consideration are actually real-valued but they are treated as ordinal ones because this provides several advantages. For instance, in \cite{pamminger2010model}, the gross wage of different individuals is divided into six ordered categories according to the quintiles of the income distribution for each year. As stated by \cite{pamminger2010model}, one of the advantages of considering wage categories relies on the fact that no inflation adjustment has to be made. Another illustrative example involves the well-known air quality index (AQI), which presents the status of daily air quality and shows the degree of air pollution in a particular place \cite{chen2021ordinal}. The air quality is often classified into six different levels which are determined according to the concentrations of several air pollutants. While the field of OTS data analysis is still in its early stages, there are already a few interesting works on the topic. In one of the first papers, \cite{bandt2005ordinal} proposed robust methods of time series analysis which use only comparisons of values and not their actual size. As previously stated, \cite{weiss2020distance} developed an interesting methodology for defining statistical features in the ordinal settings, which is based on expected distances. Later, \cite{weiss2021measuring} proposed a family of signed dependence measures for analyzing the behavior of a given OTS. There are also some recent works involving machine learning tasks in the context of ordinal series. For instance, \cite{chen2021ordinal} considered different models to forecast the air quality levels in 16 cities of Taiwan. Two novel distances between OTS were proposed in \cite{lopez2023submitted} and used to construct effective clustering algorithms. The approaches were applied to datasets of financial time series and interesting conclusions were reached. Previous references highlight the remarkable growth that OTS analysis has recently undergone.   

In accordance with previous comments, it is clear that the construction of software tools specifically designed to deal with OTS is crucial. However there exist no software packages in well-known programming languages (e.g., R \cite{rsoftware}, Python \cite{10.5555/1593511}...) aimed at dealing with ordinal series. Moreover, there are only a few libraries focusing on the analysis of ordinal data without a temporal nature, which are mostly written in the R language, but too often restricted to specific statistical procedures. For instance, the package \textbf{ordinal} \cite{ordinal} implements cumulative link models for coping with ordinal response variables. Specific functions for generating multivariate ordinal data are provided through package \textbf{MultiOrd} \cite{multiord}. In a purely machine learning context, an innovative computing tool named \textbf{ocapis} containing classification algorithms for ordinal data is described in \cite{heredia2019ocapis}. In addition, the library includes two preprocessing techniques, an instance selector and a feature selector. Note that, although their usefulness is beyond doubt, none of the previously mentioned packages is suitable to execute simple exploratory analyses, a task which should be usually performed before moving on to more sophisticated procedures. In sum, there are currently no software tools allowing to compute classical  features for ordinal series. 

The goal of this manuscript is to present th R package \textbf{otsfeatures}, which includes several functions to compute well-known statistical features for ordinal series. Besides giving valuable information about the behavior of the time series, the corresponding features can be used as input for classical machine learning procedures, as clustering, classification and outlier detection algorithms. In addition, \textbf{otsfeatures} also includes some commands allowing to perform traditional inferential tasks. The two databases of financial time series described in \cite{lopez2023submitted} are also available in the package along with a synthetic dataset containing OTS which were generated from different underlying stochastic processes. These data collections allow the users to test the commands available in \textbf{otsfeatures}. It is worth mentioning that some functions of the package can also be employed to analyze ordinal data without a temporal character.

In summary, package \textbf{otsfeatures} intends to integrate a set of simple but powerful functions for the statistical analysis of OTS into a single framework. The implementation of the package was done by using the open-source R programming language due to the role of R as the most used programming language for statistical computing. \textbf{otsfeatures} is available from the Comprehensive R Archive Network (CRAN) at \url{https://cran.r-project.org/web/packages/otsfeatures/index.html}.  

The rest of the paper is structured as follows. A summary of relevant features to analyze marginal properties and serial dependence of ordinal series is presented in Section \ref{sectionbackground}. Furthermore, some novel features measuring cross-dependence between ordinal and numerical processes are also introduced. The main functions implemented in \textbf{otsfeatures} and the available datasets are presented in Section \ref{sectionmainfunctions} after providing some background on ordinal series. In Section \ref{sectionillustration}, the functionality of the package is illustrated through several examples considering synthetic data and the financial databases included in \textbf{otsfeatures}. In addition, the process of using the output of some functions to carry out traditional data learning mining is described. Some conclusions are given in Section \ref{sectionconcludingremarks}.

\section{Analyzing marginal properties and serial dependence of ordinal time series}\label{sectionbackground}

Let $\{X_t\}_{t \in \mathbb{Z}}$, $\mathbb{Z}=\{\ldots,-1,0,1,\ldots\}$, be a strictly stationary stochastic process having the ordered categorical range $\mathcal{S}=\{s_0, \ldots, s_n\}$ with $s_0<s_1<\ldots<s_n$. The process $\{X_t\}_{t \in \mathbb{Z}}$ is often referred to as an ordinal process, while the categories in $\mathcal{S}$ are frequently called the states. Let $\{C_t\}_{t \in \mathbb{Z}}$ be the count process with range $\{0, \ldots, n\}$ generating  the ordinal process $\{X_t\}_{t \in \mathbb{Z}}$, i.e., $X_t=s_{C_t}$. It is well known that the distributional properties of $\{C_t\}_{t \in \mathbb{Z}}$ (e.g., stationarity) are properly inherited by $\{X_t\}_{t \in \mathbb{Z}}$ \cite{weiss2018introduction}. In particular, the marginal probabilities can be expressed as

\begin{equation}\label{ncp1}
	p_i=P(X_t=s_i)=P(C_t=i), \, \, i=0,\ldots,n,
\end{equation}

\noindent while the lagged joint probabilities (for a lag $l \in \mathbb{Z}$) are given by 

\begin{equation}\label{ncp2}
	p_{ij}(l)=P(X_t=s_j, X_{t-l}=s_i)=P(C_t=j, C_{t-l}=i), \, \, i,j=0,\ldots,n.
\end{equation}

\noindent Note that both the marginal and the joint probabilities are still well defined in the general case of a stationary stochastic process with nominal range, i.e., when no underlying ordering exists in $\mathcal{S}$. By contrast, in an ordinal process, one can also consider the corresponding cumulative probabilities, which are defined as

\begin{equation}\label{cp}
	\begin{split}
		f_i=P(X_t \le s_i)=P(C_t \le i), \, \, i=0,\ldots,n-1, \\
		f_{ij}(l)=P(X_t \le s_j, X_{t-l} \le s_i)=P(C_t \le j, C_{t-l}\le i), \\ i,j=0,\ldots,n-1, \, \, l \in \mathbb{Z}, 
	\end{split}
\end{equation}

\noindent for the marginal and the joint case, respectively.

In practice, the quantities $p_i$, $p_{ij}(l)$, $f_i$, and $f_{ij}(l)$ must be estimated from a $T$-length realization of the ordinal process, $\overline{X}_t=\{\overline{X}_1, \ldots, \overline{X}_T\}$, usually referred to as \textit{ordinal time series} (OTS). Natural estimates of these probabilities are given by

\begin{equation}\label{estimatesprobabilities}
	\widehat{p}_i=\frac{\sum_{k=1}^{T}I(\overline{X}_k=s_i)}{T}, \, \, \widehat{p}_{ij}(l)=\frac{\sum_{k=1}^{T-l}I(\overline{X}_k=s_i)I(\overline{X}_{k+l}=s_j)}{T-l}, 
\end{equation}
\begin{equation}\label{estimatescprobabilities}
	\widehat{f}_i=\frac{\sum_{k=1}^{T}I(\overline{X}_k \le s_i)}{T}, \, \, \widehat{f}_{ij}(l)=\frac{\sum_{k=1}^{T-l}I(\overline{X}_k \le s_i)I(\overline{X}_{k+l} \le s_j)}{T-l}, 
\end{equation}

\noindent where $I(\cdot)$ denotes the indicator function.

Probabilities $p_i$, $p_{ij}(l)$, $f_i$ and $f_{ij}(l)$ can be used to represent the process $\{X_t\}_{t \in \mathbb{Z}}$ in terms of marginal and serial dependence patterns. An alternative way of describing a given ordinal process is by means of features measuring classical statistical properties (e.g., centrality, dispersion \ldots) in the ordinal setting. A practical approach to define these quantities consists of considering expected values of some distances between ordinal categories \cite{weiss2020distance}. Specifically, a given distance measure $d: \mathcal{S}\times\mathcal{S}$ gives rise to specific ordinal features. Three of the most commonly used distances are the so-called Hamming, block and Euclidean, which are defined as

\begin{equation}
	d_\text{H}(s_i,s_j)=1-\delta_{ij}, \, \, \, d_{\text{o},1}=(s_i,s_j)=|i-j| \, \, \, \text{and} \, \, \, d_{\text{o},2}(s_i,s_j)=(i-j)^2,
\end{equation}

\noindent for a pair of states $s_i$ and $s_j$, respectively, where $\delta_{ij}$ denotes the Kronecker delta. The first six quantities in Table \ref{tablefeaturesordinalprocess} describe the marginal behavior of the process $X_t$ for a given distance $d$. There $X_t^1$ and $X_t^2$ refer to independent copies of $X_t$. In addition, the notation $X_t^r$ was used to define a reflected copy of $X_t$, that is a stochastic process independent of $X_t$ such that $P(X_t^r=s_i)=p_{n-i}$, $i=0, \ldots,n$. Note that the considered location measures pertain to the ordinal set $\mathcal{S}$. For the remaining marginal features, some assumptions are needed to obtain the ranges provided in Table \ref{tablefeaturesordinalprocess}, where $d_0^n=d(s_0, s_n)$. Particularly, for these four measures, we assume that $d_0^n=\max_{x,y \in \mathcal{S}}d(x,y)$, a property which is usually referred to as maximization. In addition, for the asymmetry, we require that $(\boldsymbol J-\boldsymbol I)\boldsymbol D$ is a positive semidefinite matrix, where $\boldsymbol{I}$ and $\boldsymbol{J}$ denote the identity and the counteridentity matrices of order $n+1$ and $\boldsymbol D=(d_{ij})_{1\leq i,j \leq n+1}$, with $d_{ij}=d(s_{i-1}, s_{j-1})$ is the corresponding pairwise distance matrix. Moreover, for both the asymmetry and the skewness to be reasonable measures, we assume that the distance $d$ is centrosymmetric, that is $d(s_i, s_j)=d(s_{n-i},s_{n-j})$, $i,j=0,\ldots,n$. Note that, for a symmetric process (that is, a process $X_t$ such that $X_t$ and $X_t^r$ have the same marginal distribution), then both the asymmetry and the skewness are expected to be zero. It is worth highlighting that, under the required assumptions, the quantities $\text{disp}_{\text{loc},d}$, $\text{disp}_{d}$, $\text{asym}_d$ and $\text{skew}_d$ can be easily normalized to the interval $[0,1]$ (or $[-1,1]$ in the case of the skewness) by dividing them by $d_0^n$. 

\begin{table}
	\centering 
	\resizebox{12cm}{!}{\begin{tabular}{cccc} \hline 
			Ordinal &  &  &  \\  
			Measure & Definition & Range & Type \\ \hline
			Location (standard) &    $x_{\mathrm{loc}, d}=\argmin _{x \in \mathcal{S}} E[d(X_t, x)]$        &   $\mathcal{S}$    &  Marginal   \\  
			Location (with respect to $s_0$) &     $x^0_{\mathrm{loc}, d}=\argmin _{x \in \mathcal{S}} |E[d(X_t, s_0)]-d(x,s_0)|$       &   $\mathcal{S}$    &  Marginal   \\  
			Dispersion (standard) &     $\text{disp}_{\text{loc},d}=E[d(X_t, x_{\mathrm{loc}, d})]$       &   $[0, d_0^n]$    &  Marginal   \\  
			Dispersion (DIVC)	& $\text{disp}_d=E[d(X_t^1, X_t^2)]$           &    $[0, d_0^n]$   &  Marginal    \\
			Asymmetry	&      $\text{asym}_d=E[d(X_t, X_t^{r})]-\text{disp}_d$      &   $[0, d_0^n]$    &   Marginal   \\
			Skewness	&     $\operatorname{skew}_d=E[d(X_t, s_n)]-E[d(X_t, s_0)]$       &   $[-d_0^n, d_0^n]$    &  Marginal    \\
			Ordinal Cohen\textquotesingle s $\kappa$ 	&     $\kappa_d(l)=\frac{\operatorname{disp}_d-E[d(X_t, X_{t-l})]}{\operatorname{disp}_d}$  & -  &  Serial   \\
			TCC &       $\Psi^c(l)=\frac{1}{r^2}\sum_{i,j=1}^{r}\psi_{ij}(l)^2$     &    $[0, 1]$   &  Serial   \\ \hline 
			Ordinal and numerical &  &  &  \\  
			Measure & Definition & Range & Type \\ \hline
			TMCLC	&    $\Psi^m_1(l)=\frac{1}{r}\sum_{i=1}^{r}\psi_i^*(l)^2$       &   $[0, 1]$    &  Serial   \\
			TMCQC &     $\Psi^m_2(l)=\frac{1}{r}\sum_{i=1}^{r}\int_{0}^{1}\psi^\rho_{i}(l)^2d\rho$       &   $[0, 1]$    &  Serial   \\ \hline 
	\end{tabular}}
	\caption{Some features of an ordinal stochastic process (top) and measuring serial dependence between an ordinal and a numerical process (bottom).}
	\label{tablefeaturesordinalprocess} 
\end{table}

Estimates of the marginal features in Table \ref{tablefeaturesordinalprocess}, denoted by means of the notation $\widehat{(\cdot)}$, where $(\cdot)$ stands for the corresponding measure (e.g., $\text{disp}_d$) can be obtained by considering estimates of $E[d(X_t, s_i)]$ ($i=0, \ldots, n$), $E[d(X_t^1, X_t^2)]$ and $E[d(X_t, X_t^r)]$ given by

\begin{equation}
	\begin{split}
		\widehat{E}[d(X_t, s_i)]=\frac{1}{T}\sum_{t=1}^{T}d(\overline{X}_t, s_i), \\
		\widehat{E}[d(X_t^1, X_t^2)]=\sum_{i,j=0}^{n}d(s_i, s_j)\widehat{p}_i\widehat{p}_j, \\ 
		\widehat{E}[d(X_t, X_t^r)]=\sum_{i,j=0}^{n}d(s_i, s_j)\widehat{p}_i\widehat{p}_{n-j},
	\end{split}
\end{equation}  

\noindent respectively. Table \ref{tablefeaturesordinalprocess} also contains some features assessing the serial dependence in an ordinal process. In this context, one of the most common quantities is the so-called ordinal Cohen\textquotesingle s $\kappa$, which measures the relative deviation of the dispersion for dependent and independent random variables at a given lag $l \in \mathbb{Z}$. This quantity can take either positive or negative values, with its upper bound being 1 and its lower bound being dependent on the underlying distance $d$. A sample version of $\kappa_d(l)$, denoted by $\widehat{\kappa}_d(l)$, is obtained by using $\widehat{\text{disp}}_d$ and the standard estimate of $E[d(X_t, X_{t-l})]$ defined as

\begin{equation}
	\widehat{E}[d(X_t, X_{t-l})]=\frac{1}{T-l} \sum_{t=l+1}^T d(\overline{X}_t, \overline{X}_{t-l}).
\end{equation}

A detailed analysis of the marginal quantities in Table \ref{tablefeaturesordinalprocess} plus the ordinal Cohen\textquotesingle s $\kappa$ is given in \cite{weiss2020distance}. In particular, the asymptotic properties of the corresponding estimates are derived and their behavior is analyzed in some simulation experiments. 

The serial dependence of an ordinal process can be evaluated by means of alternative quantities which do not pertain to the approach based on expected distances. First, let us define the cumulative binarization of the process $X_t$ as the multivariate process $\{\bm Y_t=(Y_{t,0}, \ldots, Y_{t,n-1})^\top, t \in \mathbb{Z}\}$ such that $Y_{t,i}=1$ if $X_t\leq s_i$, $i=0, \ldots, n-1$. By considering pairwise correlations in the cumulative binarization and fixing a lag $l \in \mathbb{Z}$, we obtain the quantities  

\begin{equation}\label{correlations}
	\psi_{ij}(l)=Corr(Y_{t, i}, Y_{t-l, j})=\frac{f_{ij}(l)-f_if_j}{\sqrt{f_i(1-f_i)f_j(1-f_j)}}.
\end{equation}

\noindent $i,j=0,\ldots,n-1$. The features in \ref{correlations} are very convenient because they play a similar role than the autocorrelation function of a numerical stochastic process. A measure of dependence at lag $l$ can be obtained by considering the sum of the squares of all features $\psi_{ij}(l)$. In this way, we define the total cumulative correlation (TCC) as

\begin{equation}
\Psi^c(l)=\frac{1}{n^2}\sum_{i,j=0}^{n-1}\psi_{ij}(l)^2.
\end{equation}

An estimate of the previous quantity can be obtained by considering $\widehat{\Psi}^c(l)=\frac{1}{n^2}\sum_{i,j=0}^{n-1}\widehat{\psi}_{ij}(l)^2$, where $\widehat{\psi}_{ij}(l)$ is the natural estimate of $\psi_{ij}(l)$ obtained by replacing $f_i$, $f_j$ and $f_{ij}(l)$ by $\widehat{f}_i$, $\widehat{f}_j$ and $\widehat{f}_{ij}(l)$ in \eqref{correlations} computed from the realization $\overline{X}_t$. 

Another interesting phenomenon that can be analyzed when dealing with an ordinal process is to measure the degree of cross-dependence that the process displays with respect to a given real-valued process. To this aim, let $\{Z_t, t \in \mathbb{Z}\}$ be a strictly stationary real-valued process and consider the correlation

\begin{equation}\label{totalmixedcorrelation}
	\psi_i^*(l)=Corr(Y_{t,i}, Z_{t-l})=\frac{Cov(Y_{t,i}, Z_{t-l})}{\sqrt{f_i(1-f_i)\sigma^2}},
\end{equation}

\noindent $i=1,\ldots, n-1$, which evaluates the level of linear dependence between state $s_i$ of process $X_t$ and the process $Z_t$ at a given lag $l \in \mathbb{Z}$. A more complete measure assessing general types of dependence can be constructed by defining the quantity

\begin{equation}\label{totalmixedqcorrelation}
	\psi_{i}^\rho(l)=Corr\big(Y_{t,i}, I(Z_{t-l}\leq q_{Z_t}(\rho)) \big)=\frac{Cov(Y_{t,i}, I(Z_{t-l}\leq q_{Z_t}(\rho))}{\sqrt{f_i(1-f_i)\rho(1-\rho)}},
\end{equation}

\noindent \noindent $i=1,\ldots, n-1$, where $\rho \in (0, 1)$ is a probability level, $I(\cdot)$ stands for the indicator function and $q_{Z_t}$ denotes the quantile function of process $Z_t$. Note that, by considering different values for $\rho$, dependence at different levels at lag $l$ can be evaluated between processes $X_t$ and $Z_t$.  

The features of the form $\psi_{i}^*(l)$ can be combined in a proper way to get a suitable measure of the average linear correlation between a categorical and a numerical process. In this way, we define the total mixed cumulative linear correlation (TMCLC) at lag $l$ as

\begin{equation}
	\Psi^m_1(l)=\frac{1}{n}\sum_{i=0}^{n-1}\psi^*_{i}(l)^2.
\end{equation} 

Analogously, a measure of the average quantile correlation between both processes, so-called the total mixed cumulative quantile correlation (TMCQC) at lag $l$, can be defined as

\begin{equation}
	\Psi^m_2(l)=\frac{1}{n}\sum_{i=0}^{n-1}\int_{0}^{1}\psi^\rho_{i}(l)^2d\rho.
\end{equation} 

Note that both quantities $\Psi^m_1(l)$ and $\Psi^m_2(l)$ (see the lower part of Table \ref{tablefeaturesordinalprocess}) are naturally defined in the range $[0, 1]$, with the value 0 being reached in the case of null cross-dependence between $X_t$ and $Z_t$. On the contrary, larger values indicate a stronger degree of cross-dependence between both processes. 

Natural estimates of $\Psi^m_1(l)$ and $\Psi^m_2(l)$, denoted by $\widehat{\Psi}^m_1(l)$ and $\widehat{\Psi}^m_2(l)$, respectively, can be obtained by considering standard estimates of $\psi^*_i(l)$ and $\psi^p_i(l)$, denoted by $\widehat{\psi}^*_i(l)$ and $\widehat{\psi}^p_i(l)$, respectively. To compute the latter estimates, a $T$-length realization of the bivariate process $\{(X_t, Z_t), t \in \mathbb{Z}\}$, that is $(\overline{X}_t, \overline{Z}_t) =\{(\overline{X}_1, \overline{Z}_1),\ldots, (\overline{X}_T, \overline{Z}_T)\}$, is needed. In this way, estimates $\widehat{\psi}^*_i(l)$ and $\widehat{\psi}^p_i(l)$ take the form

\begin{equation}\label{estimateswithcov}
	\begin{split}
		\widehat{\psi}^*_i(l)=\frac{\widehat{Cov}(Y_{t,i}, Z_{t-l})}{\sqrt{\widehat{f}_i(1-\widehat{f}_i)\widehat{\sigma}^2}}, \\
		\widehat{\psi}^p_i(l)=\frac{\widehat{Cov}(Y_{t,i}, I(Z_{t-l}\leq \widehat{q}_{Z_t}(\rho))}{\sqrt{\widehat{f}_i(1-\widehat{f}_i)\rho(1-\rho)}},
	\end{split}
\end{equation} 

\noindent where $\widehat{Cov}(\cdot, \cdot)$ denotes the standard estimate of the covariance between two random variables, and $\widehat{\sigma}^2$ and $\widehat{q}_{Z_t}(\cdot)$ are standard estimates of the variance and the quantile function of process $Z_t$ computed from the realization $\overline{Z}_t$. Estimates in \eqref{estimateswithcov} give rise to the quantities $\widehat{\Psi}^m_1(l)=\frac{1}{n-1}\sum_{i=1}^{n-1}\widehat{\psi}^*_{i}(l)^2$ and $\widehat{\Psi}^m_2(l)=\frac{1}{n-1}\sum_{i=1}^{n-1}\int_{0}^{1}\widehat{\psi}^\rho_{i}(l)^2d\rho$.

\section{Main functions in otsfeatures}\label{sectionmainfunctions}

This section is devoted to present the main content of package \textbf{otsfeatures}. First, the datasets available in the package are briefly described, and then the main functions of the package are introduced, including both graphical and analytical tools. 

\subsection{Available datasets in otsfeatures}\label{subsectiondatasets}

The package \textbf{otsfeatures} contains some OTS datasets which can be employed to compute ordinal features, evaluate different data mining algorithms, or simply for illustrative purposes. Specifically, \textbf{otsfeatures} includes two databases of financial time series that were introduced by \cite{lopez2023submitted}. In addition, three simulated data collections which were also used in \cite{lopez2023submitted} for the evaluation of clustering algorithms are provided. A description regarding the databases which are available in \textbf{otsfeatures} is provided below. 

\begin{itemize}
	\item \textbf{Financial datasets}. The first financial dataset contains credit ratings according to Standard \& Poors (S\&P) for the 27 countries of the European Union (EU) plus the United Kingdom \cite{weiss2020distance, lopez2023submitted}. Each country is described by means of a monthly time series with values ranging from ``D'' (worst rating) to ``AAA'' (best rating). Specifically, the whole range consists of the $n+1=23$ states $s_0, \ldots, s_{22}$, given by ``D'', ``SD'', ``R'', ``CC'', ``CCC-'', ``CCC'', ``CCC+'', ``B--'', ``B'', ``B+'', ``BB--'', ``BB'', ``BB+'', ``BBB--'', ``BBB'', ``BBB+'', ``A--'', ``A'', ``A+'', ``AA--'', ``AA'', ``AA+'' and ``AAA'', respectively. The sample period spans from January 2000 to December 2017, thus resulting serial realizations of length $T=216$. The second database consists of 9402 time series for Austrian men entering the labor market in 1975 to 1980 at an age of at most 25 years \cite{pamminger2010model}. The time series represent gross wages categories in May of successive years, which are labeled with the integers from 0 to 5. The quintiles of the income distribution for a given year were used to define the wage categories. In this way, category 0 represents individuals with the lowest incomes, while category 5 represents individuals with the highest incomes. The series exhibit individual lengths ranging from 2 to 32 years with the median length being equal to 22. Note that, as a natural ordering exists in the set of wage categories, the corresponding time series can be naturally treated as OTS. 
	\item \textbf{Synthetic datasets}. Each one of the synthetic datasets is associated with a particular ordinal model concerning the underlying count process of a given OTS, namely binomial AR($p$) \cite{weiss2009new}, binomial INARCH($p$) \cite{ristic2016binomial} and ordinal logit AR(1) (see Examples 7.4.6 and 7.4.8 in \cite{weiss2018introduction}) models for the first, second and third database, respectively. In all cases, the corresponding collection contains 80 series with $n+1=6$ categories and length $T=600$, which are split into 4 groups of 20 series each. All series in a given dataset were generated from the corresponding type of process but the coefficients of the generating model are different between groups. The specific coefficients were chosen by considering Scenarios 1, 2 and 3 in \cite{lopez2023submitted}. According to the structure of these data objects, the existence of 4 different classes can be assumed.  
\end{itemize}

It is worth highlighting that the databases available in \textbf{otsfeatures} were already considered in the literature for several purposes. Specifically, the dataset of credit ratings was employed by \cite{weiss2020distance} to perform data analysis of OTS, while the database of Austrian employees was used by \cite{pamminger2010model} to carry out clustering of categorical time series. Additionally, in \cite{lopez2023submitted}, both collections were considered for the application of clustering procedures specifically designed to deal with OTS. Thus, it is clearly beneficial for the user to have available the corresponding databases through \textbf{otsfeatures}. On the other hand, note that, in each one of the synthetic datasets, the different classes can be distinguished by means of both marginal distributions and serial dependence patterns. Hence, these data objects are suitable to evaluate the effectiveness of the features in Table \ref{tablefeaturesordinalprocess} for several machine learning problems. In fact, the usefulness of these features to carry out clustering and classification tasks (among others) in these databases is illustrated in Section \ref{subsectiondataminingtasks}. Table \ref{summarydatasets} contains a summary of the 5 datasets included in \textbf{otsfeatures}.    

\begin{table}\centering 
	\resizebox{11cm}{!}{\begin{tabular}{cccccc} \hline 
			Dataset & Object & No. Series & $T$ &$|\mathcal{S}|$ & No. Classes \\ \hline 
			Credit Ratings &    \textit{CreditRatings}       & 28  & 216 & 23 &  -     \\
			Austrian Wages	&         \textit{AustrianWages}       & 9402 &  Variable & 6 &  -       \\
			Synthetic I	&         \textit{SyntheticData1}      & 80 & 600 & 6 &  4     \\
			Synthetic II	&        \textit{SyntheticData2}   & 80 & 600 &  6 &  4        \\
			Synthetic III		&   \textit{SyntheticData3}     & 80 & 600 & 6 &   4           \\ \hline 
	\end{tabular}}
	\caption{Summary of the datasets included in \textbf{otsfeatures}.}
	\label{summarydatasets}
\end{table}

\subsection{Functions for inferential tasks}\label{subsectionfht}

In this section, we present some of the tools available in \textbf{otsfeatures} to perform classical statistical tasks. In particular, we first describe one specific plot which can be used to analyze the serial dependence structure of a given ordinal series. Afterwards, we give an overview of some functions allowing to carry out hypothesis testing and the construction of confidence intervals for the quantities introduced in the previous section. 

\subsubsection{Serial dependence plot}\label{subsubsectionserialdependenceplot}

When analyzing a real-valued processes, the autocorrelation function is a classical tool for describing the corresponding serial dependence structure. Note that, in the ordinal setting, this function can still be employed by considering the underlying count process $C_t$ introduced in Section \ref{sectionbackground}, which is indeed real-valued. However, using the autocorrelation function in this context has several drawbacks, since one is treating the ordinal process as a numerical process, thus ignoring the available information about the dissimilarity between the different ordinal categories. Therefore, an alternative, more suitable tool is required to examine the serial dependence patterns of an ordinal process. In this regard, one interesting possibility consists of considering the quantity $\kappa_d(l)$ in order to evaluate the degree of dependence exhibited by the process at a given lag $l \in \mathbb{Z}$. In fact, this quantity takes the value of 0 for an i.i.d. process, while positive or negative values are associated with different types of dependence structures. Clearly, in practice, one often works with the $T$-length realization $\overline{X}_t$ and computes the estimated Cohen\textquotesingle s $\kappa$, $\widehat{\kappa}_d(l)$, which can be used to describe the serial dependence patterns of the underlying ordinal process.

It is worth highlighting that the asymptotic distribution of, $\widehat{\kappa}_d(l)$ in the particular case of the distance $d$ being the block distance. Specifically, according to Theorem 7.2.1 in  \cite{weiss2020distance}, the distribution of the estimate $\widehat{\kappa}_{d_{\text{o}, 1}}(l)$ can be approximated by a normal distribution with mean $-\frac{1}{T}$ and variance $\frac{4}{T\widehat{\text{disp}}^2_{d_{\text{o}, 1}}}\sum_{k,l=0}^{n-1}\big(\widehat{f}_{\min\{k, l\}}-\widehat{f}_k\widehat{f}_l\big)^2$. The previous asymptotic result is rather useful in practice, since it can be used to test the null hypothesis of serial independence at lag $l$. In particular, critical values for a given significance level $\alpha$ can be computed, and these quantities do not depend on the specific lag. Thus, a serial dependence graph analogous to the ACF-based plot in the real-valued case can be constructed. Specifically, after setting a maximum lag of interest, $L$, the values of $\widehat{\kappa}_{d_{\text{o}, 1}}(l)$ for lags ranging from 1 to $L$ are simultaneously depicted in one graph. Next, the corresponding critical value is added to the plot by means of a horizontal line. According to the asymptotic approximation for $\widehat{\kappa}_{d_{\text{o}, 1}}(l)$, the critical values for an arbitrary significance level $\alpha$ are given by

\begin{equation}
\pm \frac{2\sqrt{\sum_{k,l=0}^{n-1}\big(\widehat{f}_{\min\{k, l\}}-\widehat{f}_k\widehat{f}_l\big)^2}z_{1-\alpha/2}}{\sqrt{T}\widehat{\text{disp}}_{d_{\text{o}, 1}}}-\frac{1}{T},
\end{equation}

\noindent where $z_\tau$ denotes the $\tau$-quantile of the standard normal distribution. The corresponding graph allows to easily identify the collection of significant lags for a given ordinal series. Likewise the autocorrelation plot in the numerical setting, serial dependence plots for OTS can be used for several purposes, including model selection or identification of regular patterns in the series, among others. 

The right panel of Figure \ref{2plots} shows the serial dependence plot based on $\widehat{\kappa}_{d_{\text{o}, 1}}(l)$ for the first time series in dataset \textit{AustrianWages}. A maximum lag $L=10$ was considered. The function \textit{plot\_cohens\_kappa()} was employed to construct the graph. It is worth remarking that, if the argument \textit{plot=FALSE} is used in this function, then the output is not the serial dependence plot but a list containing the values of the estimated $p$-values for the serial independence test and the critical values. 

\subsubsection{Hypothesis testing and confidence intervals}\label{subsubsectionhtaci}

Package \textbf{otsfeatures} allows to perform hypothesis tests for alternative quantities besides $\widehat{\kappa}_{d_{\text{o}, 1}}(l)$. In particular, there are some functions for testing that the quantities $\text{disp}_{d_{\text{o}, 1}}$, $\text{asym}_{d_{\text{o}, 1}}$ and $\text{skew}_{d_{\text{o}, 1}}$ are equal to some specified values employing the corresponding estimates. In addition, confidence intervals for these quantities can be constructed through some commands available in the package. In both cases, the corresponding implementations rely on the asymptotic results provided in Theorem 7.1.1 in \cite{weiss2020distance}. It is worth highlighting that these results are valid for the general case in which dependence between observations exist. However, when dealing with i.i.d. data, the corresponding expressions for the asymptotic means and variances are still valid but they get simplified. In this regard, package \textbf{otsfeatures} gives the user the possibility of performing hypothesis tests and constructing confidence intervals for i.i.d. data (see Theorem 4.1 in \cite{weiss2020distance}). This is indicated to the corresponding functions by using the argument \textit{temporal=FALSE}. 

A summary of the main functions in \textbf{otsfeatures} allowing to perform inferential tasks is given in Table \ref{tablefunctions2}.

\begin{table}\centering 
	\begin{tabular}{cc} \hline 
		Output & Function in \textbf{otsfeatures} \\ \hline 
		Serial dependence plot for $\widehat{\kappa}_{d_{\text{o}, 1}}(l)$ &    \textit{plot\_ordinal\_cohens\_kappa()}             \\
		Test based on $\widehat{\kappa}_{d_{\text{o}, 1}}(l)$ &    \textit{plot\_ordinal\_cohens\_kappa(plot = FALSE)}             \\
		Test based on $\widehat{\text{disp}}_{d_{\text{o}, 1}}$ 	&         \textit{test\_ordinal\_dispersion()}             \\
		Test based on $\widehat{\text{asym}}_{d_{\text{o}, 1}}$ 	&         \textit{test\_ordinal\_asymmetry()}             \\
		Test based on $\widehat{\text{skew}}_{d_{\text{o}, 1}}$ 	&         \textit{test\_ordinal\_skewness()}              \\ 
	    Confidence interval for ${\text{disp}}_{d_{\text{o}, 1}}$ 	&         \textit{ci\_ordinal\_dispersion()}             \\
	    Confidence interval for ${\text{asym}}_{d_{\text{o}, 1}}$ 	&         \textit{ci\_ordinal\_asymmetry()}             \\
	    Confidence interval for ${\text{skew}}_{d_{\text{o}, 1}}$ 	&         \textit{ci\_ordinal\_skewness()}              \\ \hline 
	\end{tabular}
	\caption{Some functions for inference tasks implemented in \textbf{otsfeatures}.}
	\label{tablefunctions2}
\end{table}

\subsection{Functions for feature extraction in otsfeatures}

Package \textbf{otsfeatures} contains several functions allowing to compute well-known statistical quantities for OTS measuring either marginal or serial properties. All commands of this type are based on the estimated features presented in Section \ref{sectionbackground}. A summary of the corresponding functions is given in Table \ref{tablefunctions1}. In Section \ref{subsectiondataminingtasks}, the use of several functions for feature extraction available in \textbf{otsfeatures} is illustrated through several examples. 

A summary of the main functions for feature extraction available in \textbf{otsfeatures} is provided in Table \ref{tablefunctions1}.

\begin{table}\centering 
	\resizebox{12cm}{!}{\begin{tabular}{ccccc} \hline 
			Features & Function in \textbf{otsfeatures} & & Features & Function in \textbf{otsfeatures} \\ \hline 
			$(\widehat{p}_0, \ldots, \widehat{p}_n)$	&    \textit{marginal\_probabilities()}  & & $\widehat{\text{disp}}_{d}$ &         \textit{ordinal\_dispersion\_2()}              \\
			$\big(\widehat{p}_{ij}(l)\big)_{0 \leq i, j \leq n}$	&    \textit{joint\_probabilities()}  & & $\widehat{\text{asym}}_{d}$		&      \textit{ordinal\_asymmetry()}       \\
			$(\widehat{f}_0, \ldots, \widehat{f}_{n-1})$	&    \textit{c\_marginal\_probabilities()}  & & 	$\widehat{\text{skew}}_{d}$		&         \textit{ordinal\_skewness()}           \\
			$\big(\widehat{f}_{ij}(l)\big)_{0 \leq i, j \leq n-1}$	&    \textit{c\_joint\_probabilities()}  & & $\widehat{\kappa}_d(l)$		&         \textit{ordinal\_cohens\_kappa()}              \\
			$\widehat{x}_{\mathrm{loc}, d}$	&    \textit{ordinal\_location\_1()}    & & $\widehat{\Psi}^c(l)$	&         \textit{total\_c\_cor()}           \\
			$\widehat{x}^0_{\mathrm{loc}, d}$&    \textit{ordinal\_location\_2()}   & & $\widehat{\Psi}^m_1(l)$	&         \textit{total\_mixed\_c\_cor()}         \\
			$\widehat{\text{disp}}_{\text{loc},d}$	&          \textit{ordinal\_dispersion\_1()}  & &   $\widehat{\Psi}^m_2(l)$	&      \textit{total\_mixed\_c\_qcor()}          \\ \hline 
	\end{tabular}}
	\caption{Some functions for feature extraction implemented in \textbf{otsfeatures}.}
	\label{tablefunctions1}
\end{table}

\section{Using the otsfeatures package. An illustration}\label{sectionillustration}

This section is devoted to illustrate the use of package \textbf{otsfeatures}. First we give some general considerations about the package and next, we provide some examples concerning the use of several functions for data analysis and feature extraction.   

\subsection{Some generalities about otsfeatures}

In \textbf{otsfeatures}, a $T$-length OTS with range $\mathcal{S}=\{s_0, s_1,\ldots, s_n\}$, $\overline{X}_t=\{\overline{X}_1, \ldots, \overline{X}_T\}$, is defined through a vector of length $T$ whose possible values are the integer numbers from 0 to $n$. More precisely, the realization $\overline{X}_t$ is represented by using the associated realization of the generating count process $C_t$, that is, $\overline{C}_t=\{\overline{C}_1, \ldots, \overline{C}_T\}$ such that $\overline{X}_j=s_{\overline{C}_j}$, $j=1,\ldots,T$. Note that the main advantage of this approach relies on the fact that only numerical vectors are needed for the representation of ordinal series. 

The majority of functions in the package take as input a single OTS. For instance, functions in Table \ref{tablefunctions1} return by default the corresponding estimate. Some of these functions admit the argument \textit{features=TRUE}. In that case, the function returns a vector which contains the individual quantities which are considered to construct the corresponding estimate. For instance, the function \textit{total\_c\_cor()}  computes by default the estimate $\widehat{\Psi}^c(l)$. However, if we employ the argument \textit{features=TRUE}, a matrix whose $(i,j)$ entry contains the quantity $\widehat{\psi}_{ij}(l)$ is returned. In fact, the extraction of the individual components of some estimates can be very useful for several purposes. Functions \textit{ots\_plot()} and \textit{plot\_cohens\_kappa()} with the default settings produce the corresponding time series plot and serial dependence graph, respectively. On the contrary, the remaining functions and function \textit{plot\_cohens\_kappa()} with \textit{plot=FALSE} return the results of the corresponding hypothesis tests, namely the test statistic, the critical value for a given significance level used as input and the $p$-value. It is worth remarking that most commands in \textbf{otsfeatures} require the corresponding states to be specified in a vector of the form $(0, 1, \ldots, n)$. This is done by means of the argument \textit{states}. In this way, several issues can be avoided. For instance, a particular realization may not include all the underlying ordinal values. Therefore, when analyzing such a series, one could ignore the existence of some states. This is properly solved by using the argument  \textit{states}. 

The databases included in \textbf{otsfeatures} are defined by means of a list named as indicated in the first column of Table \ref{summarydatasets}. In the case of the synthetic databases, each list contains two elements, which are described below.

\begin{itemize}
	\item The element called \textit{data} is a list of matrices with the ordinal series of the corresponding collection. 
	\item The element named \textit{classes} includes a vector of class labels associated with the objects in \textit{data}.
\end{itemize} 

On the other hand, the lists associated with datasets \textit{CreditRatings} and \textit{AustrianWages} only include the element \textit{data}, as there are no underlying class labels for these data collections. 

Let\textquotesingle s take a look at one time series in dataset \textit{AustrianWages}, which represents a specific employee of the Austrian labor market. 

\begin{verbatim}
> library(otsfeatures)
> AustrianWages$data[[10]] 
[1] 3 3 3 3 3 0 0 3 2 0 4 0 0 3 3 3 4 5 4 4 4 5
\end{verbatim}

In this series, the corresponding wage categories are identified with the integers from 0 to 5 as explained in Section \ref{subsectiondatasets} (category 0 represents the lowest incomes and category 5, the highest incomes). In this way, the previous sequence represents an individual who started with a moderate wage (category 3), then decreased his income level and finally ended up in the highest wage category. Note that this representation of the series by means of integer numbers provides a simple way of quickly examining the corresponding ordinal values.   

\subsection{Performing inferential tasks}\label{subsectioninferentialtasks}

The functions described in Section \ref{subsectionfht} allow the user to obtain valuable information from a given ordinal series. Let\textquotesingle s start by analyzing one of the time series in dataset \textit{AustrianWages}. Before carrying out inferential tasks, we are going to visualize the corresponding time series as a preliminary step. To this aim, we can employ the function \textit{ots\_plot()}, which takes as input the time series we want to represent and a vector containing the different states.

\begin{verbatim}
ots_plot(AustrianWages$data[[100]], states = 0 : 5, 
labels = 0 : 5)
\end{verbatim}

\begin{figure}[ht]
	\centering
	\includegraphics[width=0.9\textwidth]{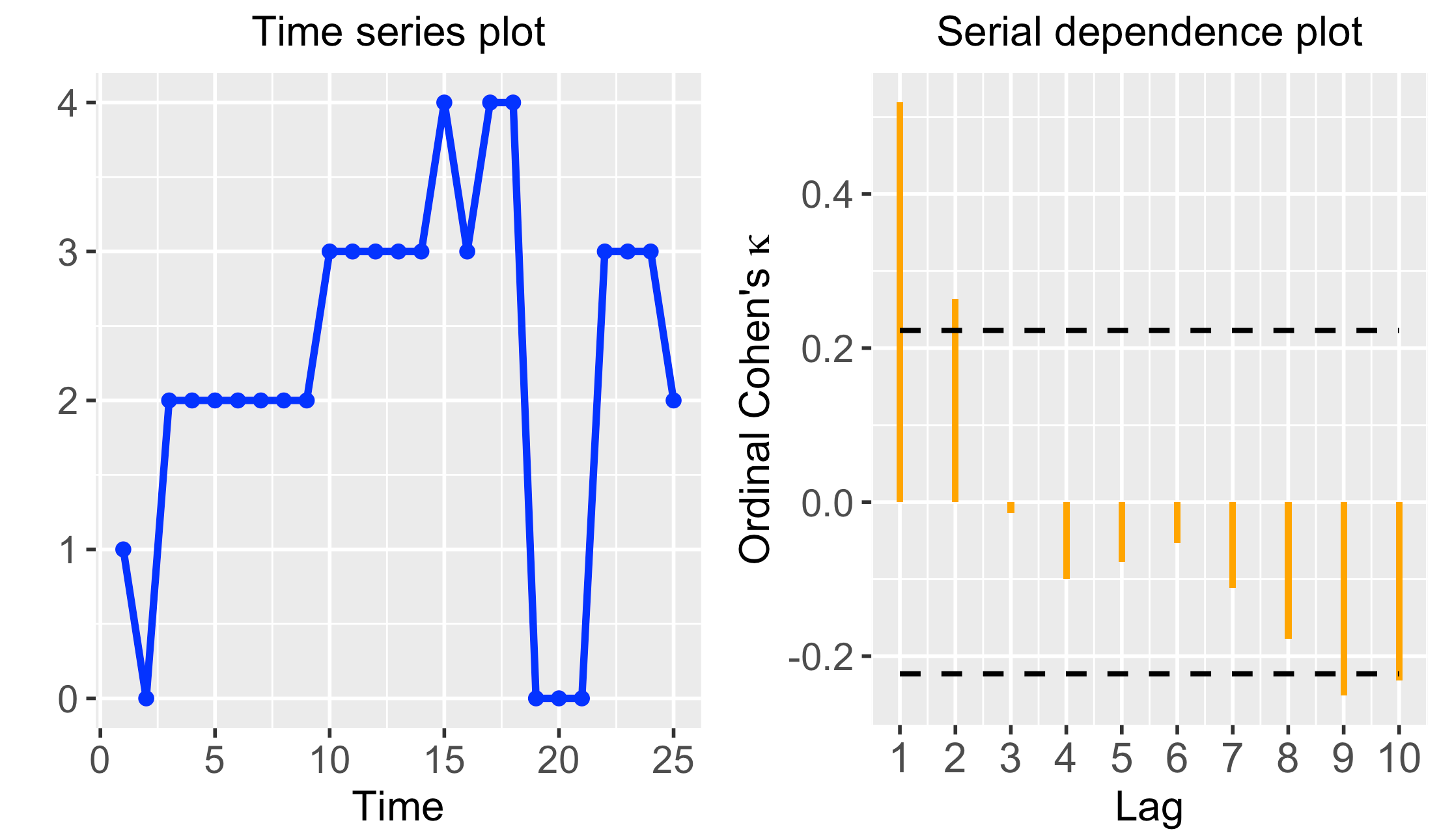}
	\caption{Time series plot (left panel) and serial dependence plot based on $\widehat{\kappa}_{d_{\text{o}, 1}}(l)$ (right panel) for one of the series in dataset \textit{AustrianWages}.}
	\label{2plots}
\end{figure}

We also employed the argument \textit{labels=0:5} to indicate that the states $s_0$ to $s_5$ are labeled with the integers from 0 to 5, since this is the  labeling used in the original dataset (see Section \ref{subsectiondatasets}). The corresponding graph is provided in the left panel of Figure \ref{2plots}. This series corresponds to an individual who belonged to all income levels except for the highest one (5). It is worth highlighting that, as the different states are located in the $y$-axis in increasing order, the plot is rather intuitive. In addition, note that, for the sake of simplicity, the different categories are treated as equidistant. Specifically, the graph is constructed by considering the block distance $d_{\text{o},1}$ between states, which is not always suitable, since the true underlying distance often depends on the specific context. Therefore, the graph produced by function \textit{ots\_plot()} should not be treated as an accurate plot of the corresponding OTS, but as a rough representation thereof. 

As stated in Section \ref{subsubsectionserialdependenceplot}, function \textit{plot\_cohens\_kappa()} in \textbf{otsfeatures} allows to construct a serial dependence plot based on the estimate $\widehat{\kappa}_{d_{\text{o}, 1}}(l)$. Let\textquotesingle s represent such plot for the series in the left panel of Figure \ref{2plots}. 

\begin{verbatim}
> sd_plot <- plot_ordinal_cohens_kappa(series = AustrianWages$data[[100]], 
states = 0 : 5)
\end{verbatim}

By default, the function considers lags from 1 to 10 (argument \textit{max\_lag}) and a significance level  $\alpha=0.05$ for the corresponding test (argument $\alpha$). The resulting graph is given in the right panel of Figure \ref{2plots}. As the standard autocorrelation plot, the corresponding estimates are displayed in a sequential order, with dashed lines indicating the critical values for the associated test. In this case, the serial dependence plot indicates significant dependence at lags 1 and 2. Moreover, dependence at lags 9 and 10 could also be considered significant, but this may be due to chance, since multiple tests are simultaneously carried out. Besides the dependence plot, function \textit{plot\_ordinal\_cohens\_kappa()} also produces numerical outputs. For instance, the corresponding $p$-values can be obtained by using the argument \textit{plot = FALSE}.

\begin{verbatim}
> sd_plot <- plot_ordinal_cohens_kappa(series = AustrianWages$data[[100]], 
states = 0 : 5, plot = FALSE)
> round(sd_plot$p_values, 2) 
[1] 0.00 0.02 0.68 0.30 0.38 0.49 0.26 0.11 0.03 0.04
\end{verbatim}

The $p$-values in the previous output corroborate that the quantity $\kappa_{d_{\text{o}, 1}}(1)$ is significantly non-null, thus confirming the existence of serial dependence at the first lag. Note that the $p$-values associated with lags 2, 9 and 10 also indicate rejection of the null hypothesis at level $\alpha=0.05$. However, the set of $p$-values should be properly adjusted to handle random rejections of the null hypothesis that can arise in a multiple testing context. For instance, the well-known Holm\textquotesingle s method, which controls the family-wise error rate at a pre-specified $\alpha$-level, could be applied to the $p$-values by executing the following command. 

\begin{verbatim}
> p.adjust(round(sd_plot$p_values, 2), method = `holm') 
[1] 0.00 0.18 1.00 1.00 1.00 1.00 1.00 0.66 0.24 0.28
\end{verbatim}

According to the corrected $p$-values, significant serial dependence still exists at lag 1, but the null hypothesis of serial independence at lags 2, 9 and 10 cannot be now rejected.

Besides analyzing serial dependence, hypothesis tests and confidence intervals for classical ordinal quantities can be constructed by using \textbf{otsfeatures} (see Section \ref{subsubsectionhtaci}). To illustrate these tasks, we consider again the previous OTS and start by testing the null hypothesis stating that the quantity $\text{skew}_{d_{\text{o}, 1}}$ is equal to 0. To this aim, we employ the function \textit{test\_ordinal\_skewness()}, whose main arguments are the corresponding ordinal series (argument \textit{series}) and the assumed value for $\text{skew}_{d_{\text{o}, 1}}$ (argument \textit{true\_skewness}), which is set to zero in this example. 

\begin{verbatim}
> test_os <- test_ordinal_skewness(series = AustrianWages$data[[100]], 
states = 0 : 5, true_skewness = 0)
> test_os$p_value
[1] 0.4239951
\end{verbatim}

The $p$-value of the test resulted 0.424. Therefore, the null hypothesis cannot be rejected at any reasonable significance level and we can assume that the series was generated from an ordinal process with 0 skewness. For illustrative purposes, let\textquotesingle s repeat the previous test by setting \textit{true\_skewness=2}.

\begin{verbatim}
> test_os <- test_ordinal_skewness(series = AustrianWages$data[[100]], 
states = 0 : 5, true_skewness = 2)
> test_os$p_value
[1] 0.02287435
\end{verbatim}

This time, the $p$-value indicates that the null hypothesis should be rejected at the standard significance level $\alpha=0.05$, that is, we could assume that the true skewness is different from 2 at that level. However, this is no longer the case for stricter significance levels (e.g., $\alpha=0.01$). 

The construction of a confidence interval for the quantity $\text{skew}_{d_{\text{o}, 1}}$ can be easily done by using the function \textit{ci\_ordinal\_skewness()}. By default, a confidence level of 0.95 is considered (argument \textit{level}). 

\begin{verbatim}
> ci_os <- ci_ordinal_skewness(series = AustrianWages$data[[100]], 
states = 0 : 5)
> ci_os  
Lower bound Upper bound
1  -0.7547583    1.794758
\end{verbatim}

The lower and upper bounds of the confidence interval are given by -0.75 and 1.79, respectively. It is worth remarking that, as we are dealing with a rather short time series ($T=25$), the interval is quite broad. In addition, note that 0 is included in the interval, which is coherent with the results of the first hypothesis test for $\text{skew}_{d_{\text{o}, 1}}$ above. Let\textquotesingle s now construct a confidence interval by considering a less strict significance level, namely 0.90. 

\begin{verbatim}
> ci_os <- ci_ordinal_skewness(series = AustrianWages$data[[100]], 
states = 0 : 5, level = 0.90)
> ci_os  
Lower bound Upper bound
1  -0.5498109    1.589811
\end{verbatim}

As expected, the new interval has a shorter length than the previous one. 

Inferential tasks for quantities ${\text{disp}}_{d_{\text{o}, 1}}$ and  ${\text{asym}}_{d_{\text{o}, 1}}$ can be carried out in an analogous way by using the corresponding functions (see Table \ref{tablefunctions2}). Moreover, these commands can also be used when dealing with i.i.d. data by using the argument \textit{temporal=FALSE}. 

\subsection{Performing data mining tasks}\label{subsectiondataminingtasks}

Jointly used with external functions, \textbf{otsfeatures} becomes a versatile and helpful tool to carry out different data mining tasks involving ordinal series. In this section, for illustrative purposes, we focus our attention on three important problems, namely classification, clustering and outlier detection. 

\subsubsection{Performing OTS classification}\label{subsectionotsclassification}

Firstly, we show how the output of the functions in Table \ref{tablefunctions1} can be used to perform feature-based classification. We illustrate this approach by considering the data collection \textit{SyntheticData1}, which contains 80 series generated from 4 different stochastic processes, each one of them giving rise to 20 OTS. The underlying processes are given by two binomial AR(1) and two binomial AR(2) models. Thus, each series in dataset SyntheticData1 has an associated class label determined by the corresponding generating process. Using the necessary functions, each series is replaced by a feature vector given by the quantities $\widehat{x}_{\mathrm{loc}, d_{\text{o}, 1}}$, $\widehat{\text{disp}}_{d_{\text{o}, 1}}$, $\widehat{\text{asym}}_{d_{\text{o}, 1}}$, $\widehat{\operatorname{skew}}_{d_{\text{o}, 1}}$, $\widehat{\kappa}_{d_{\text{o}, 1}}(1)$ and $\widehat{\kappa}_{d_{\text{o}, 1}}(2)$. In all cases, the argument \textit{distance=`Block'} (default) is used to indicate that the block distance should be employed as the underlying block distance between states. 

\begin{verbatim}
> features_1 <- unlist(lapply(SyntheticData1$data, 
ordinal_location_1, states = 0 : 5, distance = `Block'))
> features_2 <- unlist(lapply(SyntheticData1$data, 
ordinal_dispersion_2, states = 0 : 5, distance = `Block'))
> features_3 <- unlist(lapply(SyntheticData1$data, 
ordinal_asymmetry, states = 0 : 5, distance = `Block'))
> features_4 <- unlist(lapply(SyntheticData1$data, 
ordinal_skewness, states = 0 : 5, distance = `Block'))
> features_5 <- unlist(lapply(SyntheticData1$data, 
ordinal_cohens_kappa, states = 0 : 5, distance = `Block', lag = 1))
> features_6 <- unlist(lapply(SyntheticData1$data,
 ordinal_cohens_kappa, states = 0 : 5, distance = `Block', lag = 2))
> feature_dataset <- cbind(features_1, features_2, features_3, 
features_4, features_5, features_6)
\end{verbatim}

Note that the $i$th row of feature \textit{feature\_dataset} contains estimated values characterizing the marginal and serial behavior of the $i$th OTS in the dataset. Therefore, several standard classification algorithms can be applied to these matrix by means of the R package \textbf{caret} \cite{caret}. Package \textbf{caret} requires the dataset of features to be an object of class \textit{data.frame} whose last column must provide the class labels of the ele- ments and be named \textit{`Class’}. Thus, as a preliminary step, we create \textit{df\_feature\_dataset,} a version of \textit{feature\_dataset} properly arranged to be used as input to \textbf{caret} functions, by means of the following chunk of code.

\begin{verbatim}
> df_feature_dataset <- data.frame(cbind(feature_dataset,
SyntheticData1$classes)) 
> colnames(df_feature_dataset)[7] <- `Class’
> df_feature_dataset[,7] <- factor(df_feature_dataset[,7])
\end{verbatim}

The function \textit{train()} allows to fit several classifiers to the corresponding dataset, while the selected algorithm can be evaluated, for instance, by Leave-One-Out Cross-Validation (LOOCV). A grid search in the hyperparameter space of the corresponding classifier is performed by default. First we consider a standard classifier based on $k$ Nearest Neighbours ($k$NN) by using \textit{method = `knn'} as input parameter. By means of the command \textit{trControl()}, we define LOOCV as evaluation protocol.

\begin{verbatim}
> library(caret)
> train_control <- trainControl(method = `LOOCV')
> model_knn <- train(Class~., data = df_feature_dataset,
trControl = train_control, method = `knn')
\end{verbatim}

The object \textit{model\_kNN} contains the fitted model and the evaluation results, among others. The reached accuracy can be accessed as follows.
\begin{verbatim}
> max(model_knn$results$Accuracy)
[1] 0.95
\end{verbatim}

The $k$NN classifier achieves an accuracy of 0.95 in the dataset \textit{SyntheticData1}. Specifically, it produces only 4 misclassifications. Next we study the performance of the random forest and the linear discriminant analysis. To this aim, we need to set \textit{method = `rf'} and \textit{method = `lda'}, respectively. 
\begin{verbatim}
> model_rf <- train(Class~., data = df_feature_dataset, 
trControl = train_control, method = `rf')
> max(model_rf$results$Accuracy)
[1] 1

> model_lda <- train(Class~., data = df_feature_dataset, 
trControl = train_control, method = `lda')
> max(model_lda$results$Accuracy)
[1] 1
\end{verbatim}

Both approaches reach a perfect accuracy of 1, thus improving the predictive effectiveness of the random forest classifier. For illustrative purposes, let\textquotesingle s analyze the performance of the previous classifiers when the Hamming distance between ordinal categories is taken into account, which is indicated through the argument \textit{distance=`Hamming'}.  

\begin{verbatim}
> features_1 <- unlist(lapply(SyntheticData1$data, 
ordinal_location_1, states = 0 : 5, distance = `Hamming'))
> features_2 <- unlist(lapply(SyntheticData1$data, 
ordinal_dispersion_2, states = 0 : 5, distance = `Hamming'))
> features_3 <- unlist(lapply(SyntheticData1$data, 
ordinal_asymmetry, states = 0 : 5, distance = `Hamming'))
> features_4 <- unlist(lapply(SyntheticData1$data, 
ordinal_skewness, states = 0 : 5, distance = `Hamming'))
> features_5 <- unlist(lapply(SyntheticData1$data, 
ordinal_cohens_kappa, states = 0 : 5, distance = `Hamming', lag = 1))
> features_6 <- unlist(lapply(SyntheticData1$data,
ordinal_cohens_kappa, states = 0 : 5, distance = `Hamming', lag = 2))
> feature_dataset <- cbind(features_1, features_2, features_3, 
features_4, features_5, features_6)

> df_feature_dataset <- data.frame(cbind(feature_dataset,
SyntheticData1$classes)) 
> colnames(df_feature_dataset)[7] <- `Class’
> df_feature_dataset[,7] <- factor(df_feature_dataset[,7])

> model_knn <- train(Class~., data = df_feature_dataset,
trControl = train_control, method = `knn')
> max(model_knn$results$Accuracy)
[1] 0.975

> model_rf <- train(Class~., data = df_feature_dataset, 
trControl = train_control, method = `rf')
> max(model_rf$results$Accuracy)
[1] 1

> model_lda <- train(Class~., data = df_feature_dataset, 
trControl = train_control, method = `lda')
> max(model_lda$results$Accuracy)
[1] 0.975
\end{verbatim}

By considering the distance $d_{\text{H}}$, the $k$NN classifier slightly improves its performance while the linear discriminant analysis shows a small decrease in predictive effectiveness. The random forest still reaches perfect results. The classification ability of alternative sets of features, as well as the behavior of any other classifier, can be examined in an analogous way as above.

\subsubsection{Performing OTS clustering}\label{subsectionotsclustering}

The package \textbf{otsfeatures} also provides an excellent framework to carry out clustering of ordinal sequences. Let\textquotesingle s  consider now the dataset \textit{SyntheticData2} and assume that the clustering structure is governed by the similarity between underlying models. In other terms, the ground truth is given by the 4 groups involving the 20 series from the same generating process (a specific binomial INARCH($p$) process). We wish to perform clustering and, according to our criterion, the clustering effectiveness of each algorithm must be measured by comparing the experimental solution with the true partition defined by these four groups.

In cluster analysis, distances between data objects play an essential role. In our case, a suitable metric should take low values for pairs of series coming from the same stochastic process, and high values otherwise. A classical exploratory step to shed light on the quality of a particular metric consists of constructing a two-dimensional scaling (2DS) based on the corresponding pairwise distance matrix. In short, 2DS represents the pairwise distances in terms of Euclidean distances into a 2-dimensional space preserving the original values as well as possible (by minimizing a loss function). For instance, we are going to construct the 2DS for dataset \textit{SyntheticData2} by using two specific metrics between CTS proposed by \cite{lopez2023submitted} and denoted by $\widehat{d}_{1}$ and $\widehat{d}_{PMF}$. More specifically, given two OTS, $\overline{X}_t^{(1)}$ and $\overline{X}_t^{(2)}$, the distances are defined as follows.

\begin{equation}\label{distances}
\begin{split}
\widehat{d}_1\big(\overline{X}_t^{(1)}, \overline{X}_t^{(2)}\big)=& \sum_{i=0}^{n-1}\Big(\widehat{f}_i^{(1)}-\widehat{f}_i^{(2)}\Big)^2+ \sum_{k=1}^{L}\sum_{i=0}^{n-1}\sum_{j=0}^{n-1}\Big(\widehat{f}^{(1)}_{ij}(l_k)-\widehat{f}^{(2)}_{ij}(l_k)\Big)^2, \\
\widehat{d}_{PMF}\big(\overline{X}_t^{(1)}, \overline{X}_t^{(2)}\big)=& \sum_{i=0}^{n-1}\Big(\widehat{p}_i^{(1)}-\widehat{p}_i^{(2)}\Big)^2+ \sum_{k=1}^{L}\sum_{i=0}^{n-1}\sum_{j=0}^{n-1}\Big(\widehat{p}^{(1)}_{ij}(l_k)-\widehat{p}^{(2)}_{ij}(l_k)\Big)^2,
\end{split}
\end{equation}

\noindent where $L$ denotes the largest lag and superscripts (1) and (2) indicate that the corresponding estimates are based on the realizations $\overline{X}_t^{(1)}$ and $\overline{X}_t^{(2)}$, respectively. Both dissimilarities assess discrepancies between the marginal distributions (first terms) and the serial dependence structures (last terms) of both series. Therefore, they seem appropriate to group the CTS of a given collection in terms of underlying stochastic processes. However, note that the distance $\widehat{d}_1$ is based on cumulative probabilities, thus taking into account the underlying ordering existing in the series range. 

Let\textquotesingle s first create the datasets \textit{dataset\_1} and \textit{dataset\_2} with the features required to compute $\widehat{d}_{1}$ and $\widehat{d}_{PMF}$, respectively. As the series in \textit{SyntheticData2} were generated from binomial INARCH($1$) and binomial INARCH($2$) processes, we consider only the first two lags to construct the distance, i.e., we set $L=2$ (default option). We have to use the argument \textit{features = TRUE} in the corresponding functions. 

\begin{verbatim}
> list_marginal_1 <- lapply(SyntheticData2$data, 
c_marginal_probabilities, states = 0 : 5)
> list_serial_1_1 <- lapply(SyntheticData2$data, 
c_joint_probabilities, states = 0 : 5, lag = 1)
> list_serial_1_2 <- lapply(SyntheticData2$data, 
c_joint_probabilities, states = 0 : 5, lag = 2)
> dataset_marginal_1 <- matrix(unlist(list_marginal_1), 
nrow = 80, byrow = T)  
> dataset_serial_1_1 <- matrix(unlist(list_serial_1_1), 
nrow = 80, byrow = T)
> dataset_serial_1_2 <- matrix(unlist(list_serial_1_2), 
nrow = 80, byrow = T)
> dataset_1 <- cbind(dataset_marginal_1, dataset_serial_1_1, 
dataset_serial_1_2)

> list_marginal_2 <- lapply(SyntheticData2$data, 
marginal_probabilities, states = 0 : 5)
> list_serial_2_1 <- lapply(SyntheticData2$data, 
joint_probabilities, states = 0 : 5, lag = 1)
> list_serial_2_2 <- lapply(SyntheticData2$data, 
joint_probabilities, states = 0 : 5, lag = 2)
> dataset_marginal_2 <- matrix(unlist(list_marginal_2), 
nrow = 80, byrow = T)  
> dataset_serial_2_1 <- matrix(unlist(list_serial_2_1), 
nrow = 80, byrow = T)
> dataset_serial_2_2 <- matrix(unlist(list_serial_2_2), 
nrow = 80, byrow = T)
> dataset_2 <- cbind(dataset_marginal_2, dataset_serial_2_1, 
dataset_serial_2_2)
\end{verbatim}

The 2DS planes can be built using the function \textit{plot\_2d\_scaling()} of the R package \textbf{mlmts} \cite{mlmts}, which takes as input a pairwise dissimilarity matrix. 

\begin{verbatim}
> library(mlmts)
> distance_matrix_1 <- dist(dataset_1)
> plot_1 <- plot_2d_scaling(distance_matrix_1, 
cluster_labels = otsfeatures::SyntheticData2$classes)$plot
> distance_matrix_2 <- dist(dataset_2)
> plot_2 <- plot_2d_scaling(distance_matrix_2,
cluster_labels = otsfeatures::SyntheticData2$classes)$plot
\end{verbatim}

\begin{figure}[ht]
\centering
\includegraphics[width=0.8\textwidth]{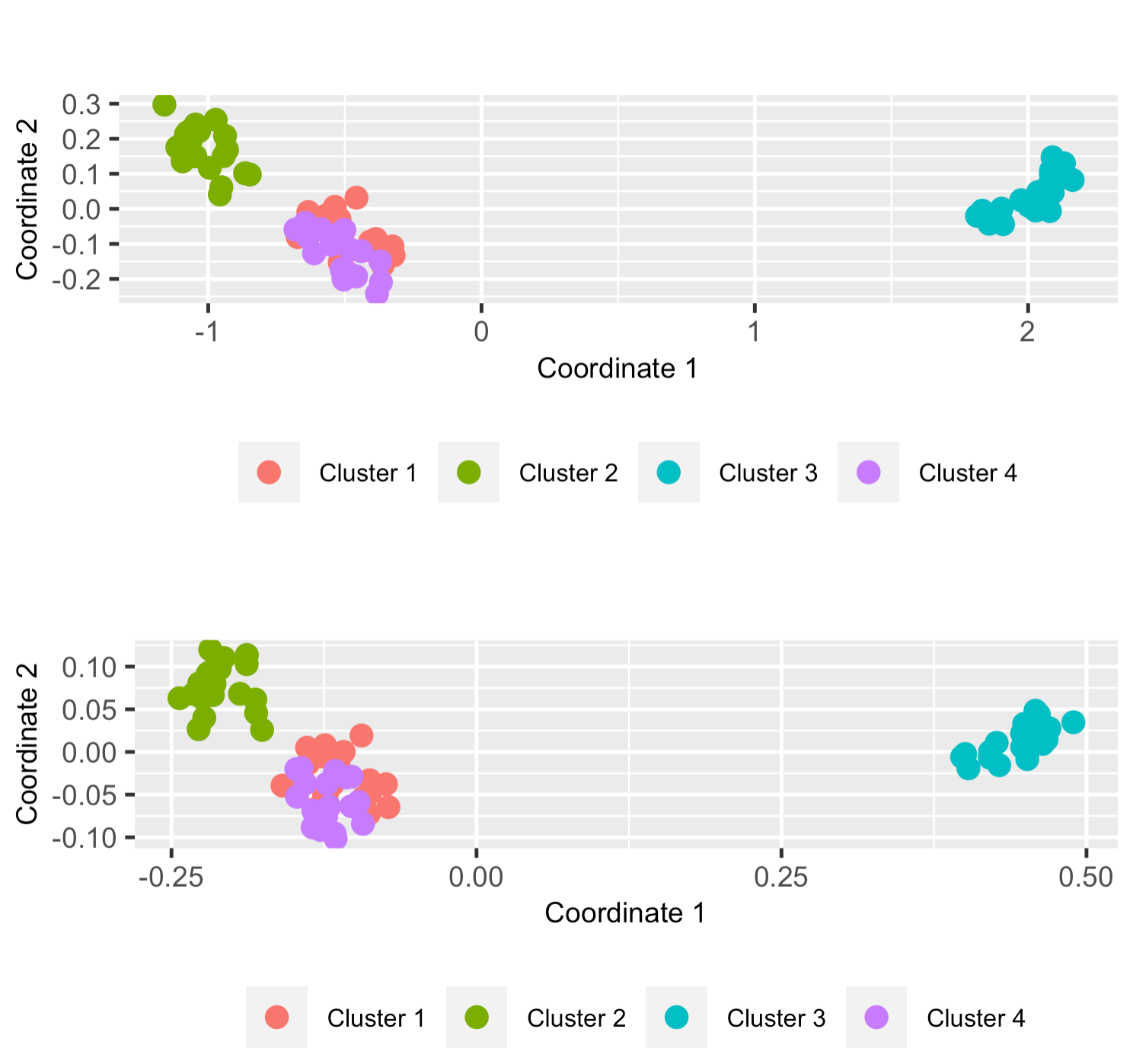}
\caption{Two-dimensional scaling planes based on distances $\widehat{d}_{1}$ (top panel) and $\widehat{d}_{PMF}$ (bottom panel) for the 80 series in the dataset \textit{SyntheticData2}.}
\label{2dplotssd1}
\end{figure}

In the previous above of code, the syntax \textit{otsfeatures::} was employed because package \textbf{mlmts} includes a data collection which is also called \textit{SyntheticData2}. The resulting plots are shown in Figure~\ref{2dplotssd1}. In both cases, the points were colored according to the true partition defined by the generating models. For it, we had to include the argument \textit{cluster\_labels} in the function \textit{plot\_2d\_scaling()}. This option is indeed useful to examine whether a specific metric is appropriate when the true class labels are known. The 2DS planes reveal that both metrics are able to identify the underlying structure rather accurately. However, there are two specific groups of OTS (the ones represented by red and purple points) exhibiting a certain degree of overlap in both plots, which suggests a high level of similarity between the corresponding generating processes. 

To evaluate the clustering accuracy of both metrics, we consider the popular Partitioning Around Medoids (PAM) algorithm, which is implemented in R through the function \textit{pam()} of package \textbf{cluster} \cite{cluster}. This function needs the pairwise distance matrix and the number of clusters. The latter argument is set to 4, since the series in dataset \textit{SyntheticData2} were generated from 4 different stochastic processes.  

\begin{verbatim}
> library(cluster)
> clustering_pam_1 <- pam(distance_matrix_1, k = 4)$clustering
> clustering_pam_2 <- pam(distance_matrix_2, k = 4)$clustering
\end{verbatim}

The vectors \textit{clustering\_pam\_1} and \textit{clustering\_pam\_2} provide the respective clustering solutions based on both metrics. The quality of both partitions requires to measure their degree of agreement with the ground truth, which can be done by using the Adjusted Rand Index (ARI) \cite{campello2007fuzzy}. This index can be easily computed by means of the function \textit{external\_validation()} of package \textbf{ClusterR} \cite{clusterr}.

\begin{verbatim}
> library(ClusterR)
> external_validation(clustering_pam_1, 
otsfeatures::SyntheticData2$classes)
[1] 0.6545303
> external_validation(clustering_pam_2, 
otsfeatures::SyntheticData2$classes)
[1] 0.6535088
\end{verbatim}

The ARI index is bounded between -1 and 1 and admits a simple interpretation: the closer it is to 1, the better is the agreement between the ground truth and the experimental solution. Moreover, the value of 0 is associated with a clustering partition picked at random according to some simple hypotheses. Therefore, it can be concluded that both metrics $d_{CC}$ and $d_B$ attain moderate scores in this dataset when used with the PAM algorithm. In particular both partitions are substantially similar. Note that a nonperfect value of ARI index was already expected from the 2DS plots in Figure \ref{2dplotssd1} due to the overlapping character of Clusters 1 and 4. 

The classical $K$-means clustering algorithm can be also executed by using \textbf{otsfeatures} utilities. In this case, we need to employ a dataset of features along with the \textit{kmeans()} function of package stats \cite{rsoftware}.

\begin{verbatim}
set.seed(123)
clustering_kmeans_1 <- kmeans(dataset_1, c = 4)$cluster
external_validation(clustering_kmeans_1, 
otsfeatures::SyntheticData2$classes)
[1] 0.6545303
> set.seed(123)
> clustering_kmeans_2 <- kmeans(dataset_2, c = 4)$cluster
> external_validation(clustering_kmeans_2, 
otsfeatures::SyntheticData2$classes)
[1] 0.7237974
\end{verbatim}

In the previous example, a slightly better results are obtained when the $\widehat{d}_{PMF}$ is employed along with the $K$-means algorithm. Concerning  $\widehat{d}_{1}$, its clustering accuracy its exactly the same as the one associated with the PAM algorithm. The performance of alternative dissimilarities or collections of features regarding a proper identification of the underlying clustering structure could be determined by following the same steps than in the previous experiments. 

\subsubsection{Performing outlier detection in OTS datasets}

Other challenging issue when analyzing a collection of OTS is to detect outlier elements. First, it is worthy noting that different notions of outlier are considered in the context of temporal data (additive outliers, innovative outliers, and others). Here, we consider the outlying elements to be whole OTS objects. More specifically, an anomalous OTS is assumed to be a series generated from a stochastic process different from those generating the majority of the series in the database. 

To illustrate how \textbf{otsfeatures} can be useful to carry out outlier identification, we create a dataset including two atypical elements. For it, we consider all the series in \textit{SyntheticData3} along with the first two series in dataset \textit{SyntheticData2}.  

\begin{verbatim}
> data_outliers <- c(SyntheticData3$data, SyntheticData2$data[1:2])
\end{verbatim}

The resulting data collection, \textit{data\_outliers}, contains 82 OTS. The first 80 OTS can be split into four homogeneous groups of 20 series, but those located into positions 81 and 82 are actually anomalous elements in the collection because they come from an ordinal logit AR(1) model (see Section~\ref{subsectiondatasets}). 

A distance-based approach to perform anomaly detection consists of obtaining the pairwise distance matrix and proceed in two steps as follows.
\begin{description}
	\item[Step 1.] For each element, compute the sum of its distances from the remaining objects in the dataset, which is expected to be large for anomalous elements. 
	\item[Step 2.] Sort the quantities computed in Step~1 in decreasing order and reorder the indexes according to this order. The first indexes in this new vector correspond to the most outlying elements, while the last ones to the least outlying elements.
\end{description}

We follow this approach to examine whether the outlying OTS in \textit{data\_outliers} can be identified by using the distance $\widehat{d}_1$ given in \eqref{distances}. First, we construct the pairwise dissimilarity matrix based on this metric for the new dataset. 

\begin{verbatim}
> list_outl_1 <- lapply(data_outliers, c_marginal_probabilities,
states = 0 : 5)
> list_outl_2 <- lapply(data_outliers, c_joint_probabilities, 
states = 0 : 5, lag = 1)
> list_outl_3 <- lapply(data_outliers, c_joint_probabilities, 
states = 0 : 5, lag = 2)
> dataset_outl_1 <- matrix(unlist(list_outl_1), nrow = 82, byrow = T)  
> dataset_outl_2 <- matrix(unlist(list_outl_2), nrow = 82, byrow = T)
> dataset_outl_3 <- matrix(unlist(list_outl_3),  nrow = 82, byrow = T)
> dataset_outl <- cbind(dataset_outl_1, dataset_outl_2, dataset_outl_3)
> distance_matrix_outl <- dist(dataset_outl)
\end{verbatim}

Then, we apply the mentioned two-step procedure to matrix \textit{distance\_matrix\_outl} by running
\begin{verbatim}
> order(colSums(as.matrix(distance_matrix_outl)), decreasing = T)[1:2]
[1] 81 82
\end{verbatim}

The previous output corroborates that $\widehat{d}_1$ is able to properly identify the two series generated from an anomalous stochastic process. As an illustrative exercise, let\textquotesingle s represent the corresponding 2DS plot for the dataset containing the two outlying OTS by using a different color for these elements.

\begin{verbatim}
> library(mlmts)
> labels <- c(otsfeatures::SyntheticData2$classes, 5, 5)
> plot_2d_scaling(distance_matrix_outl, cluster_labels = labels)$plot
\end{verbatim}

\begin{figure}[ht]
	\centering
	\includegraphics[width=0.7\textwidth]{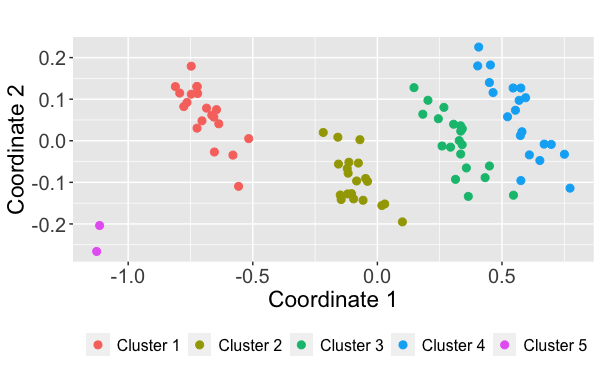}
	\caption{Two-dimensional scaling plane based on distance $\widehat{d}_1$ for the dataset containing 2 anomalous series.}
	\label{2dplotoutliers}
\end{figure}

The corresponding graph is shown in Figure \ref{2dplotoutliers}. The 2DS configuration contains four groups of points rather well-separated plus two isolated elements representing the anomalous series appearing on the left part of the plot. Clearly, 2DS plots can be very useful for outlier identification purposes, since they provide a great deal of information on both the number of potential outliers and their location with respect to the remaining elements in the dataset.

In the previous example, the number of outliers was assumed to be known, which is not realistic in practice. In fact, when dealing with real OTS databases, one usually needs to determine whether the dataset at hand contains outliers. To that aim, it is often useful to define a measure indicating the outlying nature of each object (see, e.g., \cite{weng2008detecting} and \cite{lopez2021outlier}), i.e. those  elements with an extremely large scoring could be identified as outliers. In order to illustrate this approach, we consider the dataset \textit{CreditRatings} and compute the pairwise distance matrix according to distance $\widehat{d}_{PMF}$. 

\begin{verbatim}
> list_cr_1 <- lapply(CreditRatings$data, 
marginal_probabilities, states = 0 : 22)
> list_cr_2 <- lapply(CreditRatings$data, 
joint_probabilities, states = 0 : 22, lag = 1)
> list_cr_3 <- lapply(CreditRatings$data,
joint_probabilities, states = 0 : 22, lag = 2)
> dataset_cr_1 <- matrix(unlist(list_cr_1), 
nrow = 28, byrow = T)  
> dataset_cr_2 <- matrix(unlist(list_cr_2), 
nrow = 28, byrow = T)
> dataset_cr_3 <- matrix(unlist(list_cr_3), 
nrow = 28, byrow = T)
> dataset_cr <- cbind(dataset_cr_1, dataset_cr_2, 
dataset_cr_3)
> distance_matrix_cr <- dist(dataset_cr)
\end{verbatim}

As before, the sum of the distances between each series and the remaining ones is computed. 

\begin{verbatim}
> outlier_score <- colSums(as.matrix(distance_matrix_cr))
\end{verbatim}

The vector \textit{outlier\_score} contains the sum of the distances for each one of the 28 countries. Since the $i$th element of this vector can be seen as a measure of the outlying character of the $i$th series, those countries associated with extremely large values in this vector are potential outliers. A simple way to detect these series consists of visualizing a boxplot based on the elements of \textit{outlier\_score} and checking whether there are points located into the upper part of the graph.

\begin{verbatim}
> boxplot(outlier_score, range = 1, col = `blue')
\end{verbatim}

\begin{figure}[ht]
\centering
\includegraphics[width=0.50\textwidth]{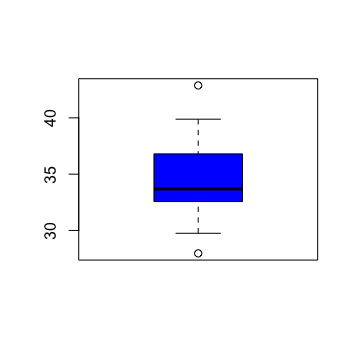}
\caption{Boxplot of the outlying scores in dataset \textit{CreditRatings} based on distance $\widehat{d}_{PMF}$.}
\label{boxplot}
\end{figure}

The resulting boxplot is shown in Figure \ref{boxplot} and suggests the existence of one series with an abnormally high outlying score. Hence, this country could be considered to be anomalous and its individual properties be carefully investigated. Specifically, the uppermost point in Figure \ref{boxplot} corresponds to Belgium. Note that the prior empirical approach provides an automatic method to determine the number of outliers. Similar analyses could be carried out by considering alternative dissimilarity measures. 

\section{Conclusions and future work}\label{sectionconcludingremarks}

Statistical analysis of time series has experienced a significant growth during the last 50 years. Although the majority of works focus on real-valued time series, ordinal time series have received a great deal of attention during the present century. The R package \textbf{otsfeatures} is fundamentally an attempt to provide different functions allowing to calculate well-known statistical quantities for ordinal series. Besides providing an useful description about the behavior of the time series, the corresponding quantities can be used as input for traditional data mining procedures, as clustering, classification and outlier detection algorithms. Additionally, \textbf{otsfeatures} includes some tools enabling the execution of classical inferential tasks, as hypothesis tests. It is worth highlighting that several functions of the package can also be used to analyze ordinal data without a temporal nature. The main motivation behind the package is that, to the best of our knowledge, no previous R packages are available for a general statistical analysis of ordinal datasets. In fact, the few software tools designed to deal with this class of databases focus on specific tasks (e.g., clustering or classification), application areas, or types of ordinal models. Package \textbf{otsfeatures} also incorporates two databases of financial time series and three synthetic datasets, which can be used for illustrative purposes. Although \textbf{otsfeatures} is rather simple, it provides the much-needed tools for the standard analyses which are usually performed before more complex tasks as modeling, inference, or forecasting. 

A description of the functions available in \textbf{otsfeatures} is given in the first part of this manuscript to make clear the details behind the software and its scope. However, the readers particularly interested in specific tools are encouraged to check the corresponding key references, which are also provided in the paper. In the second part of the manuscript, the use of the package is illustrated by considering several examples involving synthetic and real data. This can be seen as a simple overview whose goal is to make the process of using \textbf{otsfeatures} as easy as possible for first-time users.

There are three main ways through which this work can be extended. First, as \textbf{otsfeatures} is under continuous development, we expect to perform frequent updates by incorporating functions for the computation of additional statistical features which are introduced in the future. Second, note that the statistical quantities available in \textbf{otsfeatures} are defined for univariate time series. However, multivariate time series \cite{lopez2021quantile, lopez2021outlier} are becoming more and more common due to the advances in the storage capabilities of everyday devices. Thus, a software package allowing the computation of statistical features for multivariate ordinal series could be constructed in the future. First, note that \textbf{otsfeatures} assumes that all the values of a given time series are known. Although this assumption is entirely reasonable, it can become too restrictive in practice, since some OTS can include missing values. In this regard, it would be interesting to implement some functions aimed at properly correcting these values so that the computation of ordinal features does not get negatively affected.

\section*{Acknowledgments}

This research has been supported by the Ministerio de Economía y Competitividad
(MINECO) grants MTM2017-82724-R and PID2020-113578RB-100, the Xunta de Galicia (Grupos de Referencia Competitiva ED431C-2020-14), and the Centro de Investigación del Sistema Universitariode Galicia,``CITIC" grant ED431G 2019/01; all of them through the European Regional Development Fund (ERDF).

\bibliography{mybibfile}

\end{document}